\documentclass[10pt, letterpaper, conference]{IEEEtran}
\IEEEoverridecommandlockouts

\usepackage{color}

\usepackage{times}
\usepackage{epsfig}
\usepackage{graphicx}
\usepackage{tabularx}
\usepackage{amsmath}
\usepackage{amssymb}
\usepackage{bbm}
\usepackage[misc]{ifsym}
\usepackage[export]{adjustbox}
\usepackage{amsthm}
\usepackage{float}

\usepackage[ruled,vlined,linesnumbered]{algorithm2e}
\usepackage{graphicx} 

\usepackage{tikz}
\usepackage{verbatim}
\usepackage{booktabs}
\usepackage{multirow}
\usepackage{makecell}
\usepackage{authblk}
\usepackage{hyphenat} 

\usepackage{pgf}
\usepackage{algpseudocode}
\usetikzlibrary{arrows,automata}

\usepackage{graphbox}

\makeatletter
\newcommand{\removelatexerror}{\let\@latex@error\@gobble}
\makeatother

\makeatletter 
  \newcommand\figcaption{\def\@captype{figure}\caption} 
  \newcommand\tabcaption{\def\@captype{table}\caption} 
\makeatother

\makeatletter
\let\NAT@parse\undefined
\makeatother
\usepackage[colorlinks]{hyperref}  

\usepackage{graphics}
\usepackage{pifont}
\usepackage{times}

\usepackage{soul}
\usepackage{url}
\usepackage{graphicx}
\usepackage{amsmath}
\usepackage{booktabs}
\usepackage[ruled]{algorithm2e}
\usepackage{balance} 
\usepackage[T1]{fontenc}    
\usepackage{hyperref}
\usepackage{wrapfig}
\usepackage{fancybox}
\usepackage{tikz}
\usepackage{subcaption}
\usepackage{algpseudocode}
\usepackage{algorithmicx}
\usepackage[ruled,vlined,linesnumbered]{algorithm2e}
\usepackage{diagbox}
\usepackage{multirow}
\usepackage{amsfonts}       
\usepackage{nicefrac}       
\usepackage{microtype}      

\usepackage{graphbox}

\usepackage[colorlinks]{hyperref}
\graphicspath{{figures/}}

\graphicspath{{figures/}}

\begin{document}

\title{\Large \textbf{Multi-Robot Synergistic Localization in Dynamic Environments}}

\author{Ehsan Latif \and Ramviyas Parasuraman
\thanks{The authors are from the Heterogeneous Robotics Lab, Department of Computer Science, University of Georgia, Athens, GA 30602, USA. 

Email: {\it \{ehsan.latif,ramviyas\}@uga.edu}. }
}

\maketitle

\begin{abstract}
A mobile robot's precise location information is critical for navigation and task processing, especially for a multi-robot system (MRS) to collaborate and collect valuable data from the field. However, a robot in situations where it does not have access to GPS signals, such as in an environmentally controlled, indoor, or underground environment, finds it difficult to locate using its sensor alone. As a result, robots sharing their local information to improve their localization estimates benefit the entire MRS team. There have been several attempts to model-based multi-robot localization using Radio Signal Strength Indicator (RSSI) as a source to calculate bearing information. We also utilize the RSSI for wireless networks generated through the communication of multiple robots in a system and aim to localize agents with high accuracy and efficiency in a dynamic environment for shared information fusion to refine the localization estimation. This estimator structure reduces one source of measurement correlation while appropriately incorporating others. This paper proposes a decentralized Multi-robot Synergistic Localization System (MRSL) for a dense and dynamic environment. Robots update their position estimation whenever new information receives from their neighbors. When the system senses the presence of other robots in the region, it exchanges position estimates and merges the received data to improve its localization accuracy. Our approach uses Bayesian rule-based integration, which has shown to be computationally efficient and applicable to asynchronous robotics communication. We have performed extensive simulation experiments with a varying number of robots to analyze the algorithm. MRSL's localization accuracy with RSSI outperformed others on any number of robots, 66\% higher than autonomous robot localization (ARL) (which works without collaboration between robots) and 32\% higher than the collaborative multi-robot algorithm from the literature. The simulation results have shown significant promise in localization accuracy for many collaborating robots in a dynamic environment.
\end{abstract}


\section{Introduction}
\label{sec:intro}

For a robot to navigate in any area, it must be familiar with the aspects of that environment and its position concerning those features. For autonomous mobile robot systems, localization is a critical issue. In static contexts where the map does not change, current solutions to the so-called simultaneous localization and mapping (SLAM) problem \cite{grisetti2007improved} can be applied by creating a map that is then used for the rest of the localization \cite{fox2001particle,akai2022reliable}. However, while standard localization techniques work well in static environments, they may not work well in highly dynamic and complicated environments like warehouses, logistics centers, and production halls \cite{kim2016localization,ozaslan2015inspection}. The difficulty with these situations is that vast and quick changes infrequently happen, changing the map for a longer length of time. People relocating storage boxes or rearranging shelving are examples of such modifications.

Localization is an estimation of robot pose using the data collected from sensors and previously built maps. The data collected by the sensors is used by the robot to calculate its current position while in motion and to create a map of the environment. The cognition component is crucial in assessing the surroundings and determining the robot's response. A method of navigating in an already mapped or unmapped environment can be defined as the motion control of a robot or a multi-robot system. Motion control will be used to determine the robot's trajectory in various exploring tactics \cite{shue2018MRSurvey}. Moreover, multi-robot systems (MRS) can be applied to vast application scenarios such as search and rescue, warehouse logistics, etc. \cite{yang2020needs,queralta2020collaborative,han2019effective}, compared to a single robot deployment. However, determining accurate position is a challenge for a system containing multiple collaborating robots in an indoor environment with dynamic surroundings.

\begin{figure}[t]
    \centering
\begin{subfigure}{\linewidth}
\centering
\hspace{-8mm}
\includegraphics[width=0.9\linewidth]{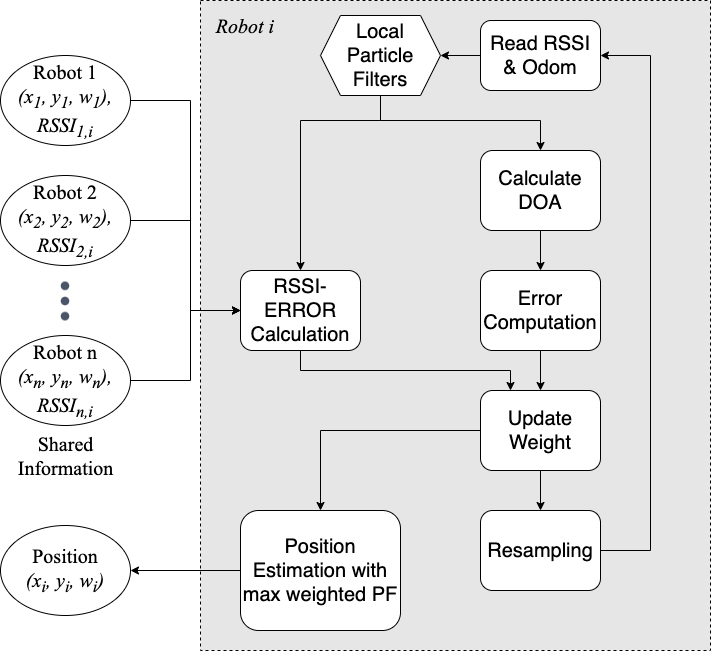}
\end{subfigure}
\caption{System architecture of the Multi-Robot Synergistic Localization (MRSL), showing input and output for robot $i$. External robot position estimates and RSSI are shown in circles out of the gray box as  robot \( j \): \((x_j, y_j, w_j)\) and \(RSSI_j\) are the best-fit pose estimates, and RSSI of robot \(j\) from its particle filter, \((x_i,y_i,w_i)\) indicates the state estimate of robot \(i\) based only on received measurements from connected \(n\) robots.}
    \label{fig:architecture}
\end{figure}
Collaborative Multi-Robot Localization (CMRL) \cite{fox1999CMRL}, which can give a drift-free map-relative position estimate for robotic robots, is one way of solving the GNSS-denied localization difficulty. CMRL compares environmental observations to a known map to determine the robot's most likely position in map coordinates.
CMRL can provide localization precision on par with map resolution in most cases. The CMRL solution's performance, on the other hand, is strongly reliant on the data available in the local topography near the robot. As a result, localization convergence may be poor or unattainable for robots crossing featureless terrain. One possible approach is for robots to work together in groups. A multi-robot strategy allows robots to share information, allowing all agents to converge even if certain people would not be able to do so otherwise.

The well-known issue of noise correlation amongst robots \cite {hamid2014navigation}\cite{prorok2011reciprocal} must be addressed by multiagent teams sharing information. The underlying assumption of most commonly used estimators, such as Extended Kalman Filters (EKF) \cite{parasuraman2018kalman} and particle filters (PF) \cite{parashar2020particle}, is that noise on new observations is uncorrelated with all past measurements. When this assumption is broken, the filter fails to capture the genuine robot state uncertainty and incorrectly converges to an inaccurate position, resulting in inconsistent estimators.

In some cases, combining data from many robots functioning in concert can reduce some problems associated with GNSS-denied navigation. For example, when faced with flat, altered, or unfamiliar terrain, a team of robots may assign several agents to stay in recognized informative zones to assist others venturing into the featureless ground. When cars navigate without external input and are vulnerable to inertial drift, combining readings from a group of robots effectively increases the total number of sensors available to each agent, reducing mistakes. In some situations, a robot may even be able to stay within GNSS signal range and communicate high-accuracy localization data with the rest of the team.

This paper proposes a decentralized synergistic localization for multi-robotic systems in a dense and dynamic environment. We propose a new Multi-Robot Synergistic Localization (MRSL) algorithm. Every robot can localize itself independently and update its position estimation using pose and sensor data from connected robots. A robot will update its position estimation whenever new information is obtained from its neighbors. When the system senses the presence of other robots in the region, it exchanges position estimates and merges the received data to improve its localization accuracy. 

For individual robot localization, we have extended the algorithm for particle filter localization \cite{parashar2020particle} of an access point in a certain way that robots localize themselves using RSSI as a source to calculate bearing as the direction of arrival (DOA). While many multi-robot collaboration algorithms from the literature use a Covariance Intersection (CI) based information integration, our approach uses Bayesian rule-based integration, which has shown to be computationally efficient and applicable to asynchronous communication. See Fig.~\ref{fig:architecture} for an overview of the MRSL.

The main contributions of this paper are outlined below.
\begin{enumerate}
   \item We propose a multi-robot synergistic localization algorithm using a particle filter updated based on the DOA and shared pose estimates with Received RSSI sensed between the robots (each robot having an AP).
   \item We verify the efficacy of the proposed MRSL algorithm through extensive numerical simulations and analyze the methods from the perspective of impacts sharing RSSI and distance information.
    \item We validate the accuracy and efficiency of the MRSL compared to autonomous robot localization (ARL) \cite{parashar2020particle}, which works without collaboration between robots, and the recent TRN-based collaborative multi-robot localization \cite{wiktor2020ICRA}, which used covariance intersection to address the temporal correlation between received signals.
\end{enumerate}

\textit{The core novelty of MRSL lies in that we employ a Bayesian rule to fuse shared information among robots in a connected network to achieve high localization accuracy}, instead of Co-variance Intersection (CI) for information merging to reduce one source of measurement correlation while appropriately incorporating others have used (as used in relevant methods in the literature). 
Our method achieves superior localization accuracy through these novelties while enabling real-time efficiency compared to several state-of-the-art solutions.

\section{Related Work}
\label{sec:relatedwork}

There has been substantial research on cooperative localization by other researchers. For example, Roumeliotis \cite{roume2002MRL} developed communal localization through the use of a decentralized Kalman filter onboard each robot in a seminal study based on the distributed heterogenous multi-robot localization \cite{madhavan2002ekf} exploited extended Kalman filter. This filter calculates robot poses while simultaneously tracking the cross-correlation terms introduced by inter-robot data sharing, which further expanded to heterogeneous robot teams and outdoor contexts, as well as close range, bearing, and orientation measurements \cite{Martinelli2005MROL}. However, these approaches are based on the Kalman filter and assume that a unimodal Gaussian may represent the posture estimate \cite{ullah2020evaluation,hamer2018clock}. These methods also need each robot's filter to estimate the condition of all other robots, which does not scale well when there are many robots.

Research work \cite{hamid2014navigation} presented a method for calculating only the robot's state utilizing inter-robot measurements to avoid assessing the status of the other robots. Through the use of Covariance Intersection (CI), the method directly addresses correlation issues. When applied to measures that are, in fact, uncorrelated, CI properly fuses measurements with an unknown amount of correlation, but it is conservative and lowers convergence. According to the authors, signals from a well-localized agent can be transferred to a robot with a poor localization estimate.
This approach, however, can only be used with an unimodal Gaussian mean and co-variance-variance representation. The Gaussian distribution utilized in a Kalman or CI filter fails to approximate position estimations in natural terrain because they are inherently multi-modal. 

Recent research has expanded on the strategies to take into account cross-correlation. To lessen its influence, \cite{Prorok2012lowcost} changed the particle resampling algorithm. However, this study used a pre-determined sampling proportion to avoid measurement correlation. Because natural terrain introduces such a wide range of correlations, relying on a pre-set heuristic does not guarantee that the filter will remain consistent. In the related subject of target tracking, \cite{7759665} employs a distributed particle filter and uses channel filters to account for cross-correlation explicitly. This strategy, however, has not been applied to the problem of localization, and channel filters are not well adapted to the intermittent communication of underwater environments since they must be reinitialized every time the network changes.

To the best of our knowledge, no previous work has met all the requirements for a robust collaborative localization technique for robots using CMRL. Therefore, we propose a particle filter-based implementation that can accommodate multi-modal CMRL estimates, explicitly accounting for cross-correlation to avoid over-convergence, with minimal computation and communication requirements which is related to the improved particle filter approach multi-sensor fusion approach \cite{khan2011fusion}. These advances allow practical implementation of the CMRL on mobile robots with robustness to network changes and communication loss.

For indoor robot localization, existing techniques can achieve high localization accuracy for wireless sensor networks (WSN) and utilizing Radio Signal Strength Indicator (RSSI) as a source to calculate bearing information \cite{liu2020survey4,owen2013haptic, morales2018low,biswas2010wifi,kotaru2015spotfi}. A performance comparison study \cite{alfurati2018performance} also backs the application of wireless sensor for localization. In collaborative multi-robot localization (CMRL), previous research used probabilistic approaches in a centralized or decentralized manner through map merging algorithms. Those techniques work well in static environments, but not in highly dynamic and complicated environments like production halls \cite{sub2019IROS}.

Approaches to addressing the localization problem in dynamic environments have a high trade-off between performance and efficiency. A recent attempt at CMRL incorporates filter architecture that allows multiple robots to collaboratively localize using Terrain Relative Navigation (TRN) \cite{wiktor2020ICRA}. For shared information fusion to refine the localization estimation, an estimator structure that uses CI to reduce one source of measurement correlation while appropriately incorporating others has been used. In contrast to current explicit modeling approaches. We aim to localize agents with high accuracy and efficiency in such an environment.

\begin{figure}[t]
\centering
 \includegraphics[width=\linewidth]{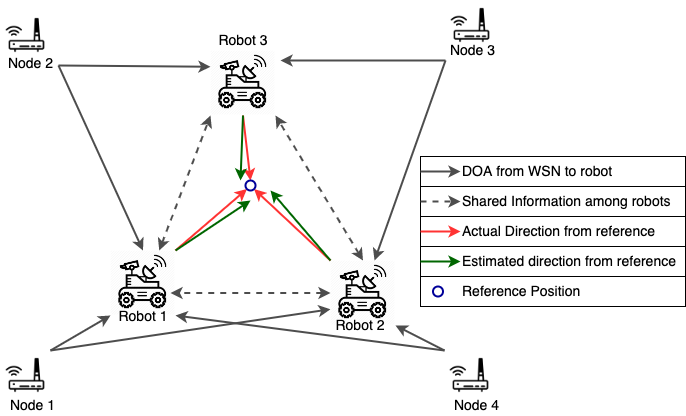}
 \caption{Overview of the proposed multi-robot synergistic localization, robots localizing themselves using local information as well as shared information.}
 \label{fig:overview}
 \vspace{-4mm}
\end{figure}

\section{Proposed Approach}
\label{sec:approach}
We look at the problem of a robot (or a wireless device) localization (self-localization) against its surroundings.
We propose a combined approach for individual and synergistic localization (see Fig.~\ref{fig:overview}).
In the individual pose estimation, several WSNs are distributed in the environment, and the mobile robot is mounted with an AP, which nearby WSNs can sense. The WSNs measure the RSSI values coming from the AP and communicate this information to the robot (assuming the measurements and shared data are reasonably time-synchronized). 
Furthermore, each robot also receives pose estimation and RSSI as shared information among the network from connected robots to improve localization accuracy. We apply the Bayesian rule to incorporate local and shared pose estimation.

Although the robot dynamics and sensors may differ, each robot's surroundings are the same, and all agents are assumed to have a synchronized clock. It is also assumed that the robots have a sensor capable of making a (potentially noisy) measurement of the position of other agents and that the robots can transmit data to other agents. This work also assumes that the agents can uniquely identify other robots and perform the associated inter-robot measurement, which is demonstrated in \cite{alexander2009CL,luo2019multi}. The robots can operate asynchronously, and communication links can be established or lost at any time.

\subsection{Individual Localization} 
Suppose that we have a wireless AP mounted on a mobile robot (whose location is to be estimated) and that four fixed WSNs \(N=\{ N_1, N_2, N_3, N_4 \}\) are placed at the corners of the bounded region (see Fig.~\ref{fig:overview} for location reference). The WSNs measure the robot's AP's radio signal strength (RSS). Using these signal values, we can estimate the DOA of the robot's vector concerning the WSN's center \cite{verma2018doa,caccamo2015extending}. The robot \(R\) records its path along with the DOA measurements as the tuple: \( m_l = \{ x_l , y_l , DOA_l\}\), where \((x_l , y_l)\) is the location of the robot at location \(l\). The problem is to find the best estimate of the robot's location \((x_{R}^*, y_{R}^* ) \in \mathbb{R}^2\) which maximizes the probability of observing the measurement tuples when the robot is at the estimated location \(P(x_{R}, y_{R} \mid m_l, m_{l-1}, . . . , m_{l-M})\), where and \(M\) is the number of previous samples considered along the completed path trajectory so far, given that we employ an arbitrary method to estimate the DOA.
 
Our solution uses a particle filter-based localization of a mobile AP for individual localization. There are two main parts of individual localization: 1) cooperative DOA estimation; 2) DOA-based localization.

\subsubsection{Cooperative Estimation of DOA} 
Cooperative localization can be accomplished with a network of three (or more) wireless nodes, where each node can sense the signal strength of the other node in the network. In our current implementation, the DOA of the mobile robot within the network is obtained from the geometric rule described in the central finite difference method \cite{parasuraman2013spatial}.
The Alg.~\ref{alg:cooperation}  gives an overview of wireless network collaboration for determining DOA.
All these computations are being performed by the moving node, which runs a centralized service and receives the RSSI information from all connected nodes that sense wireless signals independently in a synchronized manner.

RSSI can be modeled as a vector with two components, and the gradient with respect to the center of the robot can be represented as \(\vec{g} =[g_x,g_y]\). 
One of the primary advantages of the central finite difference method is that it provides gradient estimation based on the received signal strength from geometrically oriented wireless nodes. It means that, after an appropriate gradient estimation, a receiver node (moving robot) can estimate the direction of arrival of signals based on the reference position, Which then further be used for position estimation using particle filtering. 
Using the central finite difference method \cite{parasuraman2013spatial,caccamo2015extending}, the RSSI gradient can be calculated as follows:
\begin{equation}
    g_x=\frac{S_{N3} - S_{N2}}{2\Delta_{X}} + \frac{S_{N4} - S_{N1}}{2\Delta_{X}}
    \label{eqn:gx}
\end{equation}
\begin{equation}
    g_y=\frac{S_{N3} - S_{N4}}{2\Delta_{Y}} + \frac{S_{N2} - S_{N1}}{2\Delta_{Y}}
    \label{eqn:gy}
\end{equation}

Here, \(\Delta_{X}\) is the distance between the wireless sensor's antennas along the x-axis, \(\Delta_{Y}\) is the distance between the wireless sensor's antennas along the y-axis, and \(S_{N1}, S_{N2}, S_{N3}\), and \(S_{N4}\) are the RSS values at nodes \(N=\{ N_1,N_2,...N_4 \}\), respectively on the mobile robot, measured at the current path location. 
\begin{equation}
    DOA_l=\arctan(\frac{g_y}{g_x})
    \label{eqn:doa}
\end{equation}
The formula provides the DOA of the wireless signal at a position along the path \(l\) using the RSS gradient.
We can then suppress the noise of the calculated DOA by using the exponentially weighted moving normally. We consider a fixed window \(k\) of previous positions of the robot along its path. In this way, the sampled $\widetilde{DOA_l}$ at a path area \(l\) with \(K - 1\) past positions included in the window can be composed using moving average.
\begin{algorithm}[t]
 \SetAlgoLined
    \For{every time node $N_i \in N$ WSNs}{
        Each node $N_i\in N$ measures RSSI ($S_{Ni}$) locally at time $t$\;
        publish $S_{Ni}$ and $t$ to the topic $r_i$ through a publish-subscribe mechanism by all WSNs\;
    }
   \% At robot node\;
   Initialize time window to $T$\;
   \While{time window is \textit{open}}{
        \For{every RSSI topic $r_i \in R$ }{
            extract $S_{Ni}$ and $t$ from $r_i$\;
            \If{ $t$ in time window }{
                add $S_{Ni}$ to $S_i$\;
            }
            
        }
    Average out all $S_{Ni}$ in $S_i$\;
    }
   Calculate RSSI gradients: Eq.~\eqref{eqn:gx} and~\eqref{eqn:gy}\;
   Calculated DOA of robot within the WSN: Eq.~\eqref{eqn:doa}.
  \caption{Wireless sensor network (WSN) based cooperation for determination of DOA on mobile robot.}
  \label{alg:cooperation}
 \end{algorithm}

\subsubsection{DOA-based Localization}
Similar to \cite{li2015modified} for acoustic signals, we employ a Gaussian probability model on the wireless signal DOA estimates to calculate the weights of each random particle in the PF. This probabilistic model will weigh the quality of signals sensed by each node from \(N\) and ultimately produce an accurate robot location estimate through the PF. 

The error between actual DOA (ADOA) for all wireless sensors at a potential candidate position \(l\) of mobile robot with coordinates \((x_c, y_c)\) to the perceived  $\widetilde{DOA}$ values for each sensor, can be calculated as follows:
\begin{equation}
    err_l^{ws}=\sum_{j=1}^{N}(ADOA_l^j- \widetilde{DOA}_l^j) .
    \label{equation3}
\end{equation}
Now, we use the Gaussian probability formula (similar to \cite{parashar2020particle,li2015modified}) on this error to calculate the probability of the \(i_{th}\) candidate location of the particle \( q_i = (x_i, y_i, w_i), i \in (1, ..N)\), where \(N\) is the number of particles in the PF and \(w_i\) is the weight of each particle calculated over a set of previous path samples as:
\begin{equation}
    w_i  \propto P_l(q_i) = \prod_{k=0}^{M-1} \bigg[\frac{1}{\sigma\sqrt{2\pi}}e^{-\frac{(err_l^{t-k})^2}{2\sigma^2}}\bigg]
    \label{eqn:prob}
\end{equation}

There is an intrinsic angular inaccuracy in each DOA degree that is analyzed. \(\sigma\) represents this error's fluctuation (deviation), which is anticipated to be known because we know the correctness of the technique used to assess the DOA. We use the product of the Gauss likelihood of DOA error over \(M - 1\) prior robot positions (imitating geographically scattered samples) so that the sifted DOA from earlier path locations can be used in the same way as readings from many sensors. 

This component of the DOA probability is used to calculate the weights of the particles in the PF, which is then employed in the resampling procedure in the next PF step (iteration).
The particle filter provides initial hypotheses with a uniform sampling of probable robot locations across the environment using constraints around the present robot location. The Gaussian probability is determined for each particle, that is, the signal source.

 \begin{algorithm}
 \SetAlgoLined
  Pr = [] \% Initial List of Particles in the PF \;
  \While{end of trajectory stream}{
  \For{$r$ in $R$}{
  sample \(x_rt\) from transition model \(P(X_{rt+1}|x_rt)\)\;
   \(X_{rt+1} = X_rt\) \% motion model for AP embedded robot\;
   W = [] \% weights\;
   \For{\(x_{rt+1}\) in Pr}{
   add \(P(e_{rt+1}|x_{rt+1})\) to W\;
   }
   Rr = [] \% re-sampled particle filter\;
   \For{n=2 to N}{
   \For{j=1 to n}{
   Find combined probaility of $P_l(q_i)$ and $P(r_j)$ as (Eq.~\eqref{eqn:prob_j})\;
   update $w_i(x_i)$ as (Eq.~\eqref{eqn:weights})\;
   add P to Rr\;
   }
   }
   Pr = Rr\;
   \(x_l^* = x_i \in w_i(x_i)\) is \(max(w_i)\)\;
   }
  }
  \caption{Multi Robot Synergistic Localization of mobile robot using DOA from WSNs}
  \label{alg:main}
 \end{algorithm}

\subsection{Synergistic Localization}
Suppose that we have $n$ number of mobile robots, say $R = {R_1,R_2, ..., R_n}$ collaborating in the same environment. All of them are connected to each through the wireless channel, as the individual robot is equipped with a wireless transmitter and receiver.

Share information among robots $I$ contains estimated position and probability as calculated in the individual robot location process. Here, we extend the probability calculation process using Bayes's probability model by incorporating shared information as described below.

First, each robot calculate error for received and measured RSSI from neighboring robot $j$:
\begin{equation}
    err_j^{RSSI} = RSSI_m - RSSI_r
\end{equation}

where $RSSI_m$ is the measured RSSI and $RSSI_r$ is the received RSSI, we can calculate the $RSSI_m$ using path loss model.
\begin{equation} 
RSSI_m = A - 10\times n \times \log_{10}(distance(R_i,R_j))
\end{equation}
Here, \(n\) denotes the signal propagation exponent, which varies between 2 (free space) and 6 (complex indoor environment), \(d\) denotes the
distance between Robot \(R_i\) to the Robot \(R_j\), and \(A\) denotes received signal strength at a reference distance of one meter.

To get the weight for the probability based on the shared information from all neighboring robots, we can use the Gaussian probability formula same as (Eq.~\eqref{eqn:prob}) with $RSSI$ as an error.

We can further incorporate this probability $P(r_i)$ for robot $r_i$, to update the weight in Algorithm \ref{alg:main} while calculating the probability as:\\
\begin{equation}
    w_i = P_l(q_i) \times \prod_{j=1}^{n} \Big(P(r_j|x_j)\times P(x_j) \Big).
    \label{eqn:prob_j}
\end{equation}

Using a constraint around the present robot location, the particle filter provides initial hypotheses with a uniform sampling of probable robot locations across the environment. For each unique particle, that is, the signal source, the Gaussian probability is determined. The particles are subsequently given weights that are proportional to their likelihood, and the weights \(w_i\)
are normalized as:
\begin{equation}
    w_i^*(q_i)=\frac{w_i(q_i)}{\sum_{i=0}^{N-1}w_i(q_i)}.
    \label{eqn:weights}
\end{equation}
This normalized weight determines the likelihood of regenerating a set of particles in the next iteration. The particle with the highest weight (softmax) is the mobile robot's most recent best pose estimate. This process repeats until either one unique particle remains or no modern tests are available. The PF iterates for each new estimation tuple.
Algorithm.~\ref{alg:main} provides an overview of the MRSL.

The MRSL has an upper bound of complexity as \(O(nc)\) for \(c\) particle filters and \(n\) number of connected robots. Since $c$ is a constant, the algorithm runs linearly and scales linearly with the number of robots.

\begin{figure}
    \centering
    \includegraphics[width=0.9\linewidth]{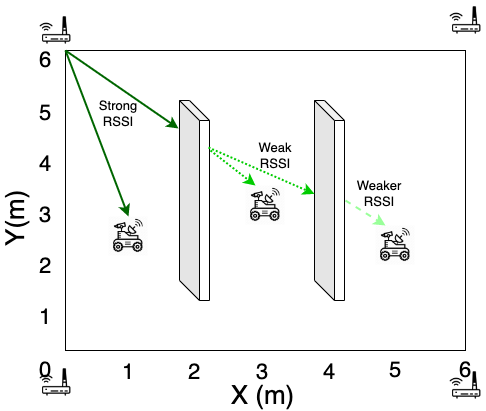}
    \caption{The simulation setup for three robots in a 6m x 6m workspace is shown here. It also depicts the two virtual walls in the simulation workspace to create non-line-of-sight conditions. Each wall will attenuate 10 dBm of RSSI signal power.}
    \label{fig:simulation-workspace}
\end{figure}

\section{Experimentation and Results}
We have performed extensive simulation experiments in a 6 x 6 meters bounded region under different scenarios to analyze the algorithm. We assumed that the infrastructure information (position of the WSN nodes) was available to all robots. The communication between connected nodes is asynchronous, i.e., the robot only updates its position estimation when it receives pose estimates from other robots. The temporal correlation among received information manages through union covariance and Bayesian rule incorporation. In all the experiments, the robots start at a random location in the workspace and move with a random walk trajectory.

Several studies have shown a range of 40 m is optimum to receive a good RSSI signal with high link quality \cite{parasuraman2013experimental,dong2012evaluation}. Therefore, we estimate that in our wireless sensor networks testbed, the nodes can be spaced with a diagonal range of up to 40 m for optimum performance in ideal conditions. This means, for a four corner node setup as in our simulation setup in Fig.~\ref{fig:simulation-workspace}, the algorithm can handle an area of 800 $m^2$. The nodes can be static or mobile, depending on the application requirements. We can increase the number of WSN nodes accordingly for a larger area. In reality, this range can be further limited due to non-line-of-sight conditions and packet drops.

We have compared our approach with the following benchmarks from the literature: 1) individual non-collaborative autonomous robot localization (ARL) with bearing (DOA) sensors \cite{parashar2020particle} that use the WSN nodes (without inter-robot collaboration)  2) Collaborative Terrain Relative Navigation (CTRN) from \cite{wiktor2020ICRA}. We implemented the proposed MRSL algorithm with the distance and RSSI sensor variants.

We have designed three sets of experiments to validate the scalability, robustness, and fault-tolerance of MRSL:
\begin{enumerate}
    \item Number of robots varying from 1 to 100; 
    \item Non-line-of-site environment with (virtual) walls generated in simulations; 
    \item Communication challenges simulated through a fixed Packet Drop Ratio (PDR). 
\end{enumerate}

Finally, we have measured the accuracy in terms of the root mean squared error (RMSE) compared to the ground truth location data from our simulations. We obtained results by performing 100 trials for each experiment set. The results shown are averaged over all these trials.

\begin{figure}[t]
\centering
\includegraphics[width=0.95\columnwidth]{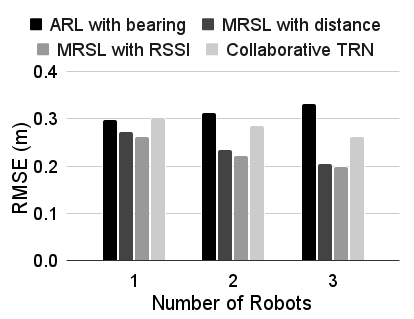}
\caption{Localization performance of different algorithms in terms of Root Mean Squared Error (RMSE) compared to the ground truth location in the 6m x 6m simulation experiments.}
\label{fig:3robotcomparison}
\end{figure}

\begin{figure}
    \centering
    \includegraphics[width=0.95\linewidth]{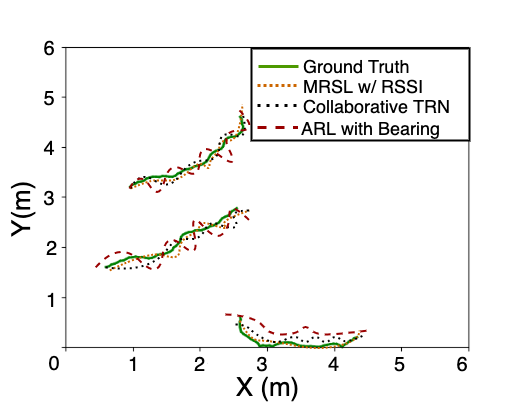}
    \caption{Trajectory of three cooperative robots in the 6m x 6m simulation experiments are shown here. The outputs of localization algorithms are plotted to comparatively observe their performance against the ground truth.}
    \label{fig:trajectory}
\end{figure}

\begin{figure*}
\centering
\begin{subfigure}{.32\linewidth}
 \includegraphics[width=\linewidth]{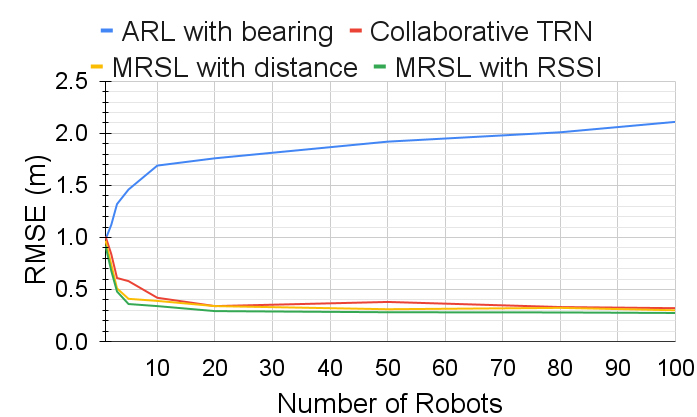}
 \caption{Number of Robots}
 \label{fig:scalability}
\end{subfigure}
\begin{subfigure}{.32\linewidth}
 \includegraphics[width=\linewidth]{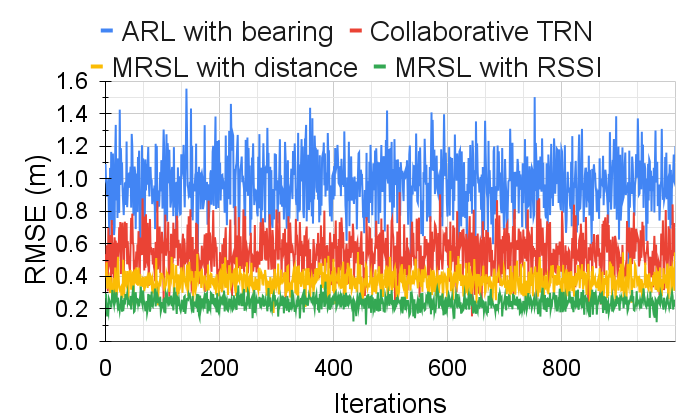}
 \caption{Non-Line-Of-Site Scenario}
 \label{fig:non-line-of-site}
\end{subfigure}
 \begin{subfigure}{.32\linewidth}
 \includegraphics[width=\linewidth]{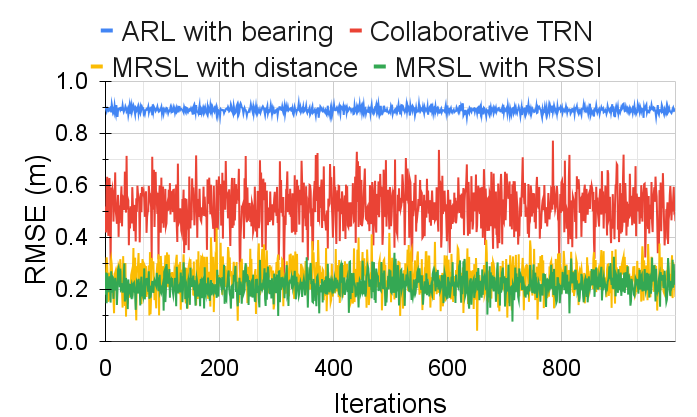}
 \caption{COM challenge: 70\% Packet Drops}
 \label{fig:com-challange}
\end{subfigure}
\caption{Comparison of different strategies to multi-robot localization (with communication of wireless signal, with communication of distance, without communication, and collaborative TRN) in terms of RMSE. \textbf{Left figure}: RMSE for increasing number of robots to validate scalability; \textbf{Center figure}: RMSE for non-line-of-site scenario with three robots to validate robustness ; \textbf{Right figure}: RMSE for communication challenge with five robots and 7-0\% PDR to validate the fault-tolerance characteristic of the proposed algorithm.}
 \label{sim-results}
 \vspace{-4mm}
\end{figure*}

\subsection{Scalability and Dense Cooperation} 

First, we obtain the level of improvement the proposed MRSL algorithm provides when the number of robots varies from one to three in the multi-robot team, with four fixed wireless sensor network anchor stations. Fig.~\ref{fig:3robotcomparison} shows the results of this experiment. It shows the superior performance of the proposed MRSL variants compared to the benchmark. Also, we can see improvement in localization accuracy with more robots in the system. This shows the effectiveness of the synergy achieved by the multi-robot team using the proposed collaborative scheme.

In Fig.~\ref{fig:trajectory}, we can see the performance of the localization in detail by comparing the trajectory generated by different multi-robot localization algorithms.

Next, we obtain more information regarding the algorithm's scalability and performance with a larger number of agents deployed with high density in a small environment. In this case, we evaluate the scalability of MRSL by increasing the number of robots from 1 to 100 with the increment of 2. Experimentation was performed on the same bounded region, which uses four WSNs distributed on the corners of the simulation workspace, with the initial position of the robots to be known with some uncertainty (Fig.~\ref{fig:overview} provides the simulation workspace setup). 

\begin{figure}[ht]
\centering
\includegraphics[width=0.95\columnwidth]{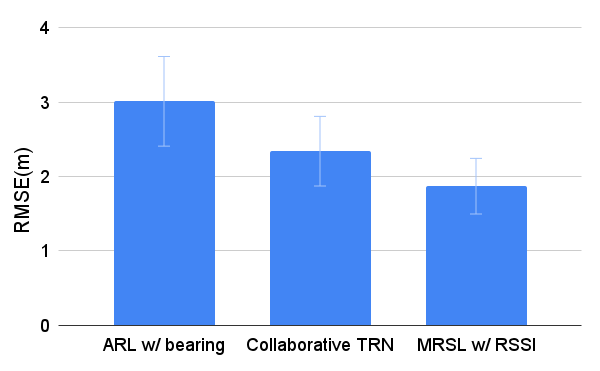}
\caption{Localization performance of different algorithms in terms of Root Mean Squared Error with three robots in the 60m x 60m larger simulation workspace.}
\label{fig:60x60}
\end{figure}

With an increasing number of robots from 1 to 10, We have observed ~35\% reduction in pose estimation error, and afterward, the error curve turned into a plateau. ARL has shown an exponential increase in error by increasing the number of the robots; however, CTRN has shown similar behaviors but less accuracy than MRSL, which can be seen in Fig.~\ref{fig:scalability}. This plateau behavior depicts the convergence for localization accuracy in multi-robotic systems and scalability with meter-level accuracy. Results also provide evidence for the sufficiency of four wireless nodes accessible to each node for multi-robot localization in a large region containing many collaborating robots.


In addition, we performed experiments with larger simulation workspace area to test the practicality of MRSL in larger environments and compare its localization accuracy with other approaches. A testbed of 60 m $\times$ 60 m has been set up for simulation with three connected moving robots with varying linear and angular velocities while considering standard dynamics and physical properties. We performed ten trials with the same simulation settings and recorded the RMSE in meters to analyze the localization accuracy. Results in Fig.~\ref{fig:60x60} have validated the claim about the practicality of MRSL for larger spaces, with a 40\%  and 21\% high localization accuracy than ARL and TRN, respectively. A standalone view of localization of MRSL is also promising, with the absolute 1.8 meters of localization error for a $3600m^2$ region, which adds more evidence for the practicality of MRSL for real-world systems.

\subsection{Dynamic Environments and Non-Line-Of-Site Conditions}
To simulate dynamic and non-line-of-sight (NLOS) environmental conditions, we added two objects in the environment: fixed (virtual) walls in the simulated region (shown in Fig.~\ref{fig:simulation-workspace}); and dynamic non-cooperative robots acting as random obstacles. Here, three non-cooperating robots executing random walks were deployed on the testbed to simulate the dynamic objects. We used three cooperating robots ($N=3$) in these experiments. The wall or robot obstacle will attenuate the RSSI a further 10 dBm if the objects come in between two cooperating robots' direct line of sight or the WSN nodes during communication.  

Fig.~\ref{fig:non-line-of-site} shows the results in the NLOS conditions. We can observe consistent meter-level accuracy even in the reduced line of sight environment for MRSL; however, ARL and CTRN ~50\% higher pose estimation error. Hence, the proposed MRSL algorithm delineates the robustness of localization in a dynamic environment.

\subsection{Robustness to Communication Challenges}
Communication difficulty is simulated through randomly dropping packets with 70\% probability, i.e., a Packet Drop Ratio (PDR) of 70\%. When packets are dropped, the robots are unable to receive information from all robots in the environment. Five robots ($N=5$) were simulated for these experiments and randomly dropped shared information packets with a fixed PDR to examine the fault tolerance of MRSL. 

Fig.~\ref{fig:com-challange} presents the results with minor fluctuation in RMSE for MRSL. Nevertheless, MRSL has been shown to be reliable while keeping the RMSE in a range. ARL was indifferent to the communication challenge, as there was no coordination among robots. However, CTRN has shown more reduction in localization accuracy than MRSL.

\section{Conclusion}
We proposed a novel multi-robot synergistic localization (MRSL) algorithm, enabling each robot in a multi-robot system to localize itself and update its position estimation using pose and sensor data from other connected robots cooperatively. In the presence of other robots in the area, it exchanges position estimates and merges the received data to improve its localization accuracy. Furthermore, we extended the algorithm for particle filter localization of an access point for individual robot localization so that robots localize themselves using the RSSI measurements as a means to calculate the inter-robot bearing using the direction of arrival (DOA) of signals. 

The proposed MRSL algorithms differ from the existing multi-robot collaboration algorithms. Instead of using the covariance intersection approach for integrating shared information, we employ Bayesian rule-based integration, which is computationally efficient. We have substantiated the practicality of MRSL in terms of scalability, robustness, and fault tolerance by performing extensive experimentation under different simulated scenarios. Nonetheless, simulation results show that the approach has much potential for use in real robots. As a result, in the future, we will deploy the MRSL on actual robots and conduct experiments under real-world conditions.

\def\bibfont{\normalfont\small}
\bibliography{ref}
\bibliographystyle{IEEEtran}

\end{document}